\pdfoutput=1

\documentclass[11pt]{article}

\usepackage[]{acl}

\usepackage{times}
\usepackage{subfigure}
\usepackage{latexsym}
\usepackage{amsmath}
\usepackage{algorithm}
\usepackage{amssymb}
\usepackage{adjustbox}
\usepackage{multirow}
\usepackage{soul}
\usepackage{color, xcolor}
\usepackage{makecell}
\usepackage{graphicx}
\usepackage{booktabs}
\usepackage{tikz-dependency}
\usepackage{pifont}
\usepackage{fdsymbol}
\usepackage{newtxtext,newtxmath}
\usepackage[cal=cm, scr=rsfs, frak=euler]{mathalfa}
\DeclareSymbolFont{extraup}{U}{zavm}{m}{n}
\DeclareMathSymbol{\varheart}{\mathalpha}{extraup}{86}
\DeclareMathSymbol{\vardiamond}{\mathalpha}{extraup}{87}

\usepackage[noend]{algpseudocode}

\definecolor{FigureOrange}{RGB}{233,112,59}
\definecolor{FigureGreen}{RGB}{83,166,60}
\definecolor{FigureBlue}{RGB}{0,159,210}
\definecolor{FigureGray}{RGB}{116,116,116}
\definecolor{EntDataset}{RGB}{255,128,0}
\definecolor{EntMethod}{RGB}{76,0,153}
\definecolor{EntTask}{RGB}{255,0,127}

\usepackage[T1]{fontenc}

\usepackage[utf8]{inputenc}

\usepackage{microtype}

\usepackage{inconsolata}

\title{SciER: An Entity and Relation Extraction Dataset for Datasets, Methods, and Tasks in Scientific Documents}

\author{\textbf{Qi Zhang}$^1$ \quad
        \textbf{Zhijia Chen}$^1$ \quad
        \textbf{Huitong Pan}$^1$ \quad \\
        \textbf{Cornelia Caragea}$^2$ \quad
        \textbf{Longin Jan Latecki}$^1$ \quad
        \textbf{Eduard Dragut}$^1$\\
  $^1$Temple University\quad $^2$University of Illinois Chicago\\ 
  {\tt \{qi.zhang, latecki, edragut\}@temple.edu,  cornelia@uic.edu} \\
  }

\begin{document}
\maketitle
\begin{abstract}
Scientific information extraction (SciIE) is critical for converting unstructured knowledge from scholarly articles into structured data (entities and relations).
Several datasets have been proposed for training and validating  SciIE models.
However, due to the high complexity and cost of annotating scientific texts, those datasets restrict their annotations to specific parts of paper, such as abstracts, resulting in the loss of diverse entity mentions and relations in context.
In this paper, we release a new entity and relation extraction dataset for entities related to datasets, methods, and tasks in scientific articles. 
Our dataset contains 106 {\em manually annotated} full-text scientific publications with over 24k entities and 12k relations.
To capture the intricate use and interactions among entities in full texts, our dataset contains 
a fine-grained tag set for relations. 
Additionally, we provide an out-of-distribution test set to offer a more realistic evaluation. 
We conduct comprehensive experiments, including state-of-the-art supervised models and our proposed LLM baselines, and highlight the challenges presented by our dataset, encouraging the development of innovative models to further the field of SciIE.\footnote{Dataset and code are publicly available: \url{https://github.com/edzq/SciER}}
\end{abstract}

\section{Introduction}
Scientific Information Extraction (SciIE) is a core topic of scientific literature mining \cite{luan-etal-2017-scientific,groth-etal-2018-open, sadat-caragea-2022-scinli, park-caragea-2023-multi, pan2024flowlearn}. It typically includes scientific named entity extraction (SciNER) and scientific relation extraction (SciRE), and
plays a critical role in downstream applications, including scientific knowledge graph construction \cite{wang-etal-2021-covid, gautam2023leveraging}, data searching \cite{viswanathan-etal-2023-datafinder}, academic question answering \cite{dasigi-etal-2021-dataset}, and method recommendation \cite{luan-etal-2018-multi}.
Scientific large language models (LLMs) like Galactica \cite{taylor2022galactica} enable several practical applications such as citations suggestion, scientific question answering (QA), and scientific code generation \cite{li2023unlocking}.
However, their generated content is frequency-biased, often exhibits overconfidence, and lacks factual basis \cite{xu-etal-2023-fine}.
SciIE, integrated with suitable retrieval, and QA systems can mitigate those issues and enhance model effectiveness in downstream tasks \cite{shu-etal-2022-tiara,xu-etal-2023-fine}.

\begin{figure}[!t]
\centering
  \resizebox{0.99\columnwidth}{!}{
    \begin{dependency}
    \begin{deptext}
   $S_1$: We train a \& \textbf{\textcolor{EntMethod}{deep CNN}} \& for \&  \textbf{\textcolor{EntTask}{semantic segmentation}}. \\ \\
    \& $E_1$: \texttt{\textcolor{EntMethod}{METHOD}} \& \& $E_2$: \texttt{\textcolor{EntTask}{TASK}} \& \\
    \end{deptext}
    \depedge[edge height=2ex]{2}{4}{USED-FOR}
    \end{dependency}
    }
 \resizebox{0.85\columnwidth}{!}{
\begin{tabular}{ccc}
\toprule
Task & Input & Output \\ \midrule
NER   & $S_1$   & $E_1$, $E_2$  \\ 
RE   & $S_1$, [$E_1$, $E_2$]  & \texttt{USED-FOR}  \\ 
ERE   & $S_1$  & [$E_1$, \texttt{USED-FOR}, $E_2$]  \\ 
\bottomrule
\end{tabular}
}
\caption{Top: An annotation sample of our SciER dataset, illustrating the labeling process and data structure. The sentence $S_1$ contains two annotated spans denoting two entities $E_1$ and $E_2$, with respective types \texttt{\textcolor{EntMethod}{METHOD}} and \texttt{\textcolor{EntTask}{TASK}}. Bottom: A table detailing the input and output of the three tasks supported by our SciER dataset, including Named Entity Recognition (NER), Relation Extraction (RE), and Entity and Relation Extraction (ERE).
}
\vspace{-15pt}
\label{fig:taste_example}
\end{figure}

SciIE faces unique challenges 
compared to general domain IE.
First, data annotation for SciIE is highly dependent on expert annotators, resulting in a scarcity of high-quality labeled datasets.
Second, 
SciIE needs to handle more complex text, which evolves constantly with novel terminology, unlike general domain IE. 
For instance, SciIE faces more severe temporal and conceptual shifts \cite{zhang2019invest,viswanathan-etal-2021-citationie, zaporojets2022tempel, chen2022web, chen2024comquest, pham-etal-2023-solving}, whereas fundamental entities and relationships in general IE tend to remain more static over time compared to those in the scientific literature.

Existing SciIE datasets and benchmarks that support both SciNER and SciRE are limited to extracting information from specific parts of papers, such as particular paragraphs \cite{augenstein-etal-2017-semeval} or abstracts \cite{gabor-etal-2018-semeval,luan-etal-2018-multi}. 
However, scientific entities like datasets, methods, and tasks entities, are distributed throughout the entire text of papers.
Sentences in the body of a paper exhibit diverse linguistic styles and ways to mention entities
\cite{ li2023unlocking} and semantics \cite{jain-etal-2020-scirex}, which allows the extraction of more fine-grained and precise relation types. For example, abstracts do not say that method X is trained on dataset Y, but experimental sections give such details.
Therefore, focusing on specific parts of scientific articles 
is likely to miss important information. 
Several datasets \cite{pan-etal-2024-scidmt-large, pan-etal-2023-dmdd,otto-etal-2023-gsap, jain-etal-2020-scirex} attempt to create SciIE benchmarks with full-text annotation, but they ignore the SciRE task.


In this paper, we present SciER, an entity and relation extraction dataset for identifying dataset, method, and task entities in scientific documents as well as the relations between them. 
Our dataset is large, with 24K entities and 12k relations from 106 scientific articles, enabling 
the evaluation and development of SciIE models.
These documents are taken from the publications included in Papers with Code (PwC)\footnote{\url{https://paperswithcode.com/}}, 
covering artificial intelligence (AI) topics, such as natural language processing (NLP), machine learning (ML), computer vision (CV), and AI for Science (AI4Science).
Figure \ref{fig:taste_example} shows an annotated sentence from our dataset, which gives the entities, their types, i.e., \texttt{METHOD} and \texttt{TASK}, respectively, and the relation between them \texttt{USED-FOR}.
Our dataset can be used to evaluate NER and RE as separate tasks, but it can also support the evaluation of end-to-end entity and relation extraction (ERE) from scientific publications \cite{luan-etal-2018-multi, ye-etal-2022-packed}.
The table in Figure \ref{fig:taste_example} describes those settings. For example, in NER the input is a sentence and the output is the set of entities in the sentence. In RE, the input is the sentence along with the entities and the output is the relation between those entities. Finally, in ERE the triplet <\textit{subject}, \textit{relation}, \textit{object}> is the expected output from a sentence.

We address the limitations of existing datasets by 
annotating entire scientific papers for both entity and their relations. This is a much harder task compared to annotating abstracts.
Furthermore, comparing with existing datasets \cite{augenstein-etal-2017-semeval,gabor-etal-2018-semeval,luan-etal-2018-multi}, we provide more fine-grained relation types to describe the interactions between datasets, methods, and tasks.
For example, we use \texttt{TRAINED-WITH} and \texttt{EVALUATED-WITH} to describe the interactions between methods and datasets. These relation types need to be extracted from the body of a paper, and are not supported by previous datasets. 
\S\ref{sec:dataset_stat_compare} gives a detailed comparison between our dataset and existing ones.
Finally, to evaluate the model's robustness to temporal and conceptual shifts in the SciIE, we set in-distributed (ID) and out-of-distribution (OOD) test sets.
The documents in the OOD set were all published after the training documents and feature entirely different topics.
We conduct evaluation experiments by employing three state-of-the-art supervised methods and LLMs-based in-context learning (ICL) methods and provide analysis.
Specifically, for LLMs-based methods, we tested both pipeline and joint approaches, optimizing the prompts through retrieval-based ICL, tag-based entity extraction, and the incorporation of annotation guidelines. The experimental results show that for LLMs, pipeline modeling, which splits the ERE task into two sub-tasks of NER and RE, outperformas joint extraction.
In the challenging ERE task, the best supervised method achieves an F1 score of 61.10\%, while the best LLM method achieves an F1 score of 41.22\%. 

Our contributions can be summarized as follows:
\vspace{-6pt}
\begin{itemize}
    \vspace{-6pt}
    \item We provide a manually annotated dataset consisting of 106 full-text scientific publications, containing over 24k entities and 12k relations. Our dataset is significantly larger than previous datasets that support both SciNER and SciRE tasks.
    \vspace{-6pt}
    \item We introduce a fine-grained tag set designed for scientific relation extraction, customized to reflect the use and interaction of machine learning datasets, methods, and tasks entities in scientific publications. 
    \vspace{-6pt}
    \item We conducted  experiments on LLMs baselines using both pipeline and joint approaches. We optimized the prompt through retrieval-based ICL, tag-based entity extraction, and the incorporation of annotation guidelines. We also provided a comparative analysis between LLMs methods and three state-of-the-art supervised baselines, highlighting the key challenges.
\end{itemize}

\section{Related Work} \label{sec:related_work}
\setlength{\tabcolsep}{3pt}
\begin{table}[t]
    \centering
    \resizebox{1\linewidth}{!}{
    \begin{small}
    \begin{tabular}{lcccc}
    \toprule
                     & SemEva17         & SemEval18 & SciERC   & SciER    \\ \hline
    Annotation Unit        &  \ding{168} & \ding{169}   & \ding{169} & \ding{171} \\
    \#Entity Types   & 3                  & -          & 6        & 3         \\
    \#Relation Types & 2                  & 6          & 7        & 9         \\
    \#Entities       & 9946               & 7483       & 8089     & 24518     \\
    \#Relations      & 672                & 1595       & 4716     & 12083     \\
    \#Docs           & 500                & 500        & 500      & 106       \\
    \#Relations/Doc  & 1.3                & 3.2        & 9.4      & 114.0     \\ 
    \bottomrule
    \end{tabular}
    \end{small}
    }
    \vspace{-10pt}
    \caption{Comparison of SciER and 3  datasets supporting NER and RE in scientific text. Annotation units: \ding{168}=Paragraph, \ding{169}=Abstract, \ding{171}=Full Text.}
    \label{tab: data_compare}
\end{table}
\setlength{\tabcolsep}{6pt}

Many datasets for SciNER have been proposed.
\cite{heddes2021automatic} and DMDD \cite{pan-etal-2023-dmdd} are two datasets for dataset mention detection. The \cite{heddes2021automatic} dataset comprises 6000 annotated sentences selected based on the occurrence of dataset related word patterns from four major AI conference publications.
DMDD is annotated on the full text and comprises 31219 scientific articles automatically annotated with distant supervision \cite{zhang-etal-2018-regular}.
TDMSci \cite{hou-etal-2021-tdmsci} 
supports three types of entities: \texttt{TASK}, \texttt{DATASET}, and \texttt{METHOD}. It has 2000 sentences extracted from NLP papers.
SciREX \cite{jain-etal-2020-scirex} offers comprehensive coverage with 438 full text annotated documents and supports four entity types: \texttt{TASK}, \texttt{DATASET}, \texttt{METHOD}, and \texttt{METRIC}. SciREX does not annotate relations between pairs of those entity types.
\cite{otto-etal-2023-gsap} manually annotates 100 documents for fine-grained SciNER by defining 10 different entity types in 3 categories: MLModel related, Dataset related and miscellaneous. 
SciDMT \cite{pan-etal-2024-scidmt-large} uses the PwC as knowledge created a very large scale dataset for \texttt{DATA}, \texttt{METHOD}, and \texttt{TASK}. SciDMT includes 48 thousand scientific articles with over 1.8 million weakly annotated mention annotations in their main corpus. However, given the inherent complexity of the NER task, employing weak labels may cause models to overfit on noisy data, thereby substantially impacting their performance \cite{liu-etal-2021-noisy-labeled, bhowmick2022boosting, bhowmick2023globally}.

Although there has been growing interest in research on developing methods and datasets for SciIE,
very few datasets support both NER and RE tasks for scientific text. 
An overview of existing SciIE benchmarks that support both SciNER and SciRE is shown in Table \ref{tab: data_compare}.
SEMEVAL-2017 TASK 10 (SemEval 17) \cite{augenstein-etal-2017-semeval} includes 500 paragraphs from open-access journals and supports three types of entities: \texttt{TASK}, \texttt{METHOD}, and \texttt{MATERIAL} and two relation types: \texttt{HYPONYM-OF} and \texttt{SYNONYM-OF}. 
SEMEVAL-2018 TASK 7 (SemEval 18) \cite{gabor-etal-2018-semeval} has been proposed for predicting six types of relations between entities.
All sentences in SemEval 18 are from the abstracts of NLP papers and have only entity spans (i.e., without annotation of entity types).
SciERC \cite{luan-etal-2018-multi} contains 500 scientific abstracts with the annotations for scientific entities, their relations, and coreference clusters. SciERC defines six types of entities and seven types of relations.
However, these three datasets are limited on annotating abstracts or pre-selected paragraphs.
Thus, a significant number of sentences that contain more diverse entity mention forms and semantics are lost.

Compared to those resources, our dataset contains 106 scientific publications with minute manual annotations.
The dataset has nine relation types, allowing for more nuanced relations between entities.
The scale of our dataset, which contains more than 24k entities and over 12k relations, which is significantly larger than previous datasets, except for those that are created with distant supervision.

\section{SciER} \label{sec:Dataset}
In this section,  we detail the curation of our dataset, including data collection process in \S\ref{sec:collection}, the data annotation process in \S\ref{sec:annotation}, and present the final dataset statistics and comparisons in \S\ref{sec:dataset_stat_compare}.

\subsection{Data Collection and Processing} \label{sec:collection}
Our dataset includes 106 documents from two sources. 
\ding{182} One hundred of these documents come from the SciDMT validation set (SciDMT-E). These documents are from the  PwC website and we use the corresponding PDF parsed version released by the S2ORC \cite{lo-etal-2020-s2orc}. These papers cover different machine learning topics and have publication dates prior to 2022. We re-check the entity annotations from SciDMT-E\footnote{We provide details of our re-checking workload on SciDMT-E in the Appendix \ref{apx:SciDMT_reann}.} and then add relation annotations.
\ding{183} We selected additional six papers from top AI conferences as an out-of-distribution (OOD) test set. To simulate a more realistic application scenario, we chose these six papers published in 2023-2024, four of which focus on AI4Science topics not included in the first 100 documents. For these six OOD test documents, we first collected their PDF files and then used Grobid \cite{GROBID} for parsing.
\subsection{Data Annotation}\label{sec:annotation}

\begin{table*}[ht]
\resizebox{\textwidth}{!}{
\centering
\begin{tabular}{llc}
\toprule
Relation Type & Explanation & Example \\
\hline
\texttt{EVALUATED-WITH}  &  Methods are evaluated by datasets &
\begin{dependency}
\begin{deptext}
We use \& \textbf{\textcolor{EntDataset}{COCO}} \& to evaluate \&  \textbf{\textcolor{EntMethod}{ConerNet-Lite}} \& and compare it wither other detectors. \\
\end{deptext}
\depedge[edge height=1ex]{4}{2}{EVALUATED-WITH}
\end{dependency}
\\
\hline
\texttt{COMPARE-WITH} & Entities are linked by comparison relation &
\multirow{2}{*}{
\begin{dependency}
\begin{deptext}
 \& \textbf{\textcolor{EntMethod}{MAC}} \& ...outperforms all tested  \&  \textbf{\textcolor{EntMethod}{RANSAC-fashion estimators}} \&, such as \& \textbf{\textcolor{EntMethod}{SAC-COT}} \&... \\
\end{deptext}
\depedge[edge height=1ex]{2}{4}{COMPARE-WITH}
\depedge[edge height=1ex]{6}{4}{SUBCLASS-OF}
\end{dependency}
}
\\
\cline{1-2}
\texttt{SUBCLASS-OF} &  One method is a specialized class of another   & \\
\hline
\texttt{BENCHMARK-FOR} & Datasets are used to evaluate tasks &
\multirow{3}{*}{
\begin{dependency}
\begin{deptext}
 \& \textbf{\textcolor{EntDataset}{FlyingChairs}} \& is a synthetic dataset designed for training \&  \textbf{\textcolor{EntMethod}{CNNs}} \& to \& \textbf{\textcolor{EntTask}{estimate optical flow}} \&. \\
\end{deptext}
\depedge[edge height=1ex]{4}{2}{TRAINED-WITH}
\depedge[edge height=3ex]{2}{6}{BENCHMARK-FOR}
\depedge[edge height=1ex]{4}{6}{USED-FOR}
\end{dependency}
}
\\
\cline{1-2}
\texttt{TRAINED-WITH} &  Methods are trained by datasets   & \\
\cline{1-2}
\texttt{USED-FOR} &  Entities are linked by usage relation   & \\
\hline
\texttt{SUBTASK-OF}
 &  A specific part of another broader Task & 
\begin{dependency}
\begin{deptext}
...is critical for \& \textbf{\textcolor{EntTask}{dense prediction tasks}} \& such as \&  \textbf{\textcolor{EntTask}{object detection}} \& ...  \\
\end{deptext}
\depedge[edge height=1ex]{4}{2}{SUBTASK-OF}
\end{dependency}
\\
\hline
\texttt{PART-OF}
 &  Entities are in a part-whole relation & 
\begin{dependency}
\begin{deptext}
Adding \& \textbf{\textcolor{EntMethod}{attention}} \& to our \&  \textbf{\textcolor{EntTask}{deep learning-based network}} \&  translated to... \\
\end{deptext}
\depedge[edge height=1ex]{4}{2}{PART-OF}
\end{dependency}
\\
\hline
\texttt{SYNONYM-OF}  &  Entities have same or very similar meanings &
\begin{dependency}
\begin{deptext}
 ...to improve \& \textbf{\textcolor{EntMethod}{Generative Adversarial Network}} \& ( \& \textbf{\textcolor{EntMethod}{GAN}} \& ) for ... 
 \\
\end{deptext}
\depedge[edge height=1ex]{4}{2}{SYNONYM-OF}
\end{dependency}
\\
\bottomrule
\end{tabular}
}
\caption{\label{tab:rel_typology}
Semantic relation typology for \texttt{\textcolor{EntDataset}{DATASET}}, \texttt{\textcolor{EntMethod}{METHOD}}, and \texttt{\textcolor{EntTask}{TASK}} entities.
}
\vspace{-15pt}
\end{table*}

\noindent\textbf{Annotation Scheme}\quad For the entity annotation, we use the SciDMT annotation scheme, which defined three types of entities: \texttt{DATASET}, \texttt{METHOD}, and \texttt{TASK}. 
To maintain consistency with the PwC website database, we only annotate the factual entities, unlike previous works \cite{luan-etal-2018-multi,otto-etal-2023-gsap} which annotate both factual and non-factual entities. 
For example, the ``CoNLL03'' and ``SNLI'' are factual entities, but the ``a high-coverage sense-annotated corpus'' is not a factual entity.

For the relation annotation, we define nine fine-grained tag set to establish interaction relationships between datasets, methods, and tasks entities in scientific documents. 
They are \texttt{EVALUATED-WITH}, \texttt{COMPARE-WITH}, \texttt{SUBCLASS-OF}, \texttt{BENCHMARK-FOR}, \texttt{TRAINED-WITH}, \texttt{USED-FOR}, \texttt{SUBTASK-OF}, \texttt{PART-OF}, and \texttt{SYNONYM-OF}.
Directionality is taken into account except for the two symmetric relation types (\texttt{SYNONYM-OF} and \texttt{COMPARE-WITH}).
We provide our semantic relation typology and corresponding examples in Table \ref{tab:rel_typology}. 
Specifically, 
compared to previous datasets \cite{augenstein-etal-2017-semeval,luan-etal-2018-multi,gabor-etal-2018-semeval}, we employ more specific relation types for identical entity types and extend usage relations among different types of entities in a more granular manner.
For example, 
we use \texttt{SUBTASK-OF} and \texttt{SUBCLASS-OF} to describe the hierarchical relations between tasks and methods, respectively. 
This can provide better interpretability and allows for direct usage in practical applications such as building taxonomies.
Additionally, we use \texttt{TRAINED-WITH} and \texttt{EVALUATED-WITH} to describe the more precise interactions between methods and datasets.
We provide more detailed definitions of the labels for entities and relations in our annotation guidelines in Appendix \ref{apx:guideline}.

\noindent\textbf{Annotation Strategy}\quad We have five annotators with backgrounds in computer science and machine learning. 
We conduct the annotation using INCEpTION\footnote{\url{https://inception-project.github.io/}} platform.
All annotators had annotation training before starting to annotate on assigned documents.
For the 100 documents from SciDMT-E, we asked annotators to first re-check the SciNER annotation before proceeding to the SciRE annotation. For the six OOD documents, annotators need to annotate both SciNER and SciRE from scratch.

\noindent\textbf{Human Agreement}\quad One annotator leads the entire annotation process and annotates all the documents in the dataset and each document is also annotated by at least two other annotators.
For the first 100 documents, the kappa score \cite{davies1982measuring} for entity annotation is 94.2\%, relation annotation is 70.8\%; for the six OOD documents, the kappa score for entity annotation is 74.1\%, relation annotation is 73.8\%.
The almost perfect agreement of entity annotation on the first 100 documents is because we derive the original annotation from SciDMT-E.

\subsection{Dataset Statistics and Comparison} \label{sec:dataset_stat_compare}

After the annotation process, our dataset contains over 24k entities and 12k relations, with each document averaging about 114 relations. As shown in Table \ref{tab: data_compare}, our dataset is significantly larger than previous datasets supporting both entity and relation extraction task. Specifically, for the widely used SciERC dataset, when we only consider Dataset, Method, and Task entities, it contains only about 1.5k entity and 1.5k relation annotations, where more details are provided in Appendix \ref{apx:scierc_compare}.
We randomly split the first 100 documents into train, development, and ID test sets, containing 80, 10, and 10 documents, respectively.
We used six OOD documents as the OOD test set.
Appendix \ref{apx: dataset_stat} lists the number of samples for each relation type in each set of our dataset.

\section{Experiments} \label{sec:Experiments}
In this section, we provide the details of evaluation experiments of both state-of-the-art supervised baselines and LLMs-based baselines on the proposed dataset. 
We first formally define the problem of end-to-end relation extraction in \S\ref{sec:problem}, 
then describe the supervised methods in  \S\ref{sec:supervised} and the LLMs-based methods in \S\ref{sec:llm}.
Finally, we present our implementation details in \S\ref{sec:implementation} and evaluation settings in \S\ref{sec:evaluation}.

\subsection{Problem Definition}\label{sec:problem}
We aim our dataset as a means to train and evaluate SciIE models.
Formally, the input document is denoted as $D$, which contains a sequence of paragraphs ${P}=\{{p}_{1}, {p}_2, \dots, {p}_n\}$. 
Each paragraph ${p}$ is composed of a sequence of sentences $\{{s}_1, {s}_2, \dots {s}_n\}$ and each sentence is composed of a sequence of words $\{{w}_1, {w}_2,  \dots, {w}_n\}$. 
Formally, the problem of end-to-end relation extraction can be decomposed into two sub-tasks:

\noindent\textbf{Named Entity Recognition}\quad Let $\mathbb{E}$ denote a set of pre-defined entity types. The NER task is to identify all entity  mentions from the input sentence ${s} =\{{w}_1, {w}_2, ..., {w}_n\}$. For each identified entity, we need to give its span ${e}_{i} = \{{w}_{l}, ..., {w}_{r}\}$, where $l$ and $r$ represent the left and right word indices of the span, and classify its entity type $t\in\mathbb{E}$.

\noindent\textbf{Relation Extraction}\quad Let $\mathbb{R}$ denote a set of pre-defined relation types. The task is to predict the relation type ${r}\in \mathbb{R}$ for every pair of entities $({e}_{i}, {e}_{j})$, if one exists,  and ${r}=\{\textrm{NULL}\}$ otherwise. 
Since end-to-end relation extraction comprises two subtasks, this task is typically addressed using \ding{182} joint entity and relation extraction (ERE) or \ding{183} pipeline extraction, i.e., performing the NER task first and then using the NER results for RE.

\subsection{Supervised Baselines} \label{sec:supervised}
We apply three supervised methods: \ding{182} \textbf{PURE} \cite{zhong-chen-2021-frustratingly} utilizes two independent encoders to perform pipeline extraction. The outputs of entity encoder are fed into the relation encoder to facilitate end-to-end relation extraction. 
This method emphasizes the significance of unique representations for entities and relations, the early integration of entity information, and leveraging global context to improve performance.
\ding{183} \textbf{PL-Marker} \cite{ye-etal-2022-packed} introduces a novel span representation technique that augments the outputs of pre-trained encoders to perform pipeline extraction. It leverages two specialized packing strategies—neighborhood-oriented for identifying entity boundaries, and subject-oriented for classifying complex span pairs—which helps understand the interrelations between spans.
\ding{184} \textbf{HGERE} \cite{yan-etal-2023-joint} proposes a joint ERE method by incorporating a high-recall pruner to reduce error propagation and by employing a hypergraph neural network to model complex interactions among entities and relations. This approach has led to significant performance improvements, establishing new state-of-the-art results in the joint ERE.

\subsection{LLMs-based Baselines} \label{sec:llm}
\begin{figure*}[!h]
     \centering
     \begin{subfigure}
         \centering
         \includegraphics[width=0.24\textwidth]{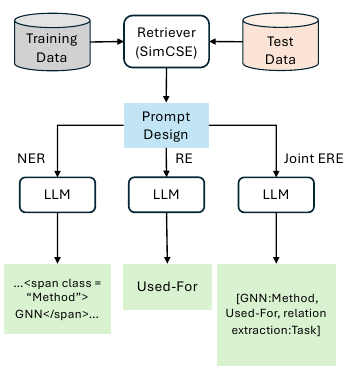}
         \label{fig:few_shot}
     \end{subfigure}
     \begin{subfigure}
         \centering
         \includegraphics[width=0.24\textwidth]{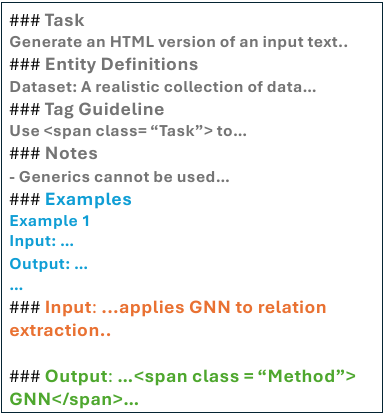}
         \label{fig:ner_prompt}
     \end{subfigure}
     \begin{subfigure}
         \centering
         \includegraphics[width=0.24\textwidth]{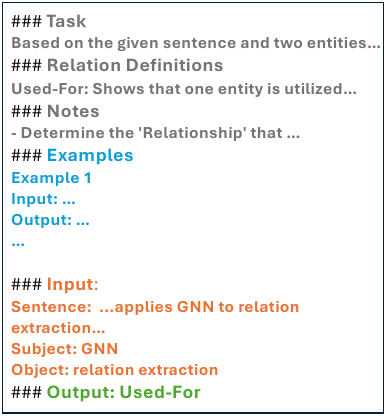}
         \label{fig:re_prompt}
     \end{subfigure}
     \begin{subfigure}
         \centering
         \includegraphics[width=0.24\textwidth]{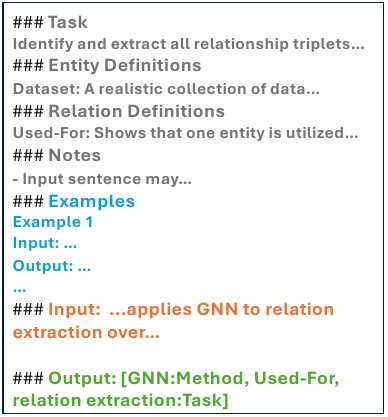}
         \label{fig:e2e_prompt}
     \end{subfigure}
    \vspace{-16pt}
    \caption{
    Overall architecture of LLM in-context learning (few-shot) baselines for NER, RE and joint Entity and Relation Extraction (ERE) (first). The few-shot prompt templates for NER (second), RE (third), and Joint ERE (fourth). Different colors indicate different prompt design elements: \textcolor{FigureGray}{gray} for annotation guideline-based task instructions ${I}$, \textcolor{FigureBlue}{blue} for retrieved demonstrations  ${D}$, \textcolor{FigureOrange}{orange} denotes the test example input $x_{test}$, and the \textcolor{FigureGreen}{green} represents the expected output of test example output, which will be omitted during testing. $y_{test}$. Due to space constraints, we shortened the text of our prompts.}
    \label{fig:overall_system_and_examples}
\vspace{-17pt}
\end{figure*}

LLMs via in-context learning (ICL) represents a significant advancement in NLP \cite{qin-etal-2023-chatgpt}.   
To comprehensively evaluate the LLMs' capability on SciIE, we employ LLMs with zero-shot and few-shot settings to perform both pipeline extraction and joint ERE.
Several studies  suggest that choosing few-shot
in-context examples for each test example dynamically instead of using a fixed set of in-context examples yields strong improvements for ICL \cite{jimenez-gutierrez-etal-2022-thinking, liu-etal-2022-makes}. In our experiments, we follow this setting by employing a retriever to find top similar samples from training set as in-context examples. We will first detail the prompt template construction to formalize the NER, RE, and joint ERE as a language generation task \cite{jimenez-gutierrez-etal-2022-thinking}. 
Then we will introduce the specific settings and efforts to improve the prompt for each task.

We construct an unique prompt for each given test example, which is fed to the LLM. Each prompt consists of the following components:

\noindent\textbf{Instruction} ${I}$\quad The task instruction ${I}$ provides the LLMs with a basic description of the task the LLM needs to perform and in what format it should output the results.

\noindent\textbf{Demonstrations} ${D}$\quad The demonstrations are retrieved from the training set for  as in-context examples to help the model better understand the task. Specifically, we will employ a retriever to compute the sentence similarity score and acquire the most similar $k$ demonstrations $(x_i, y_i)$ to build ${D}$.

\noindent\textbf{Test Input} $x_{test}$\quad Following the same format as demonstrations, we offer the test input $x_{test}$, and LLM is expected to generate the corresponding output result $y_{test}$.

In summary, LLMs-based few-shot in-context learning (ICL) for each task can be formulated as:
\begin{equation} \label{eq:prompt}
    P(y_{test}|{I}, {D}, x_{test})
\end{equation}
When performing zero-shot ICL, ${D}$ will be removed from the prompt.

Our LLMs-based baseline framework is shown in Figure \ref{fig:overall_system_and_examples}.
Existing work indicates that for information extraction tasks, LLMs require clearer instructions to improve the performance\cite{qin-etal-2023-chatgpt,hu2024improving,sainz2023gollie,jimenez-gutierrez-etal-2022-thinking}. 
Therefore, we use annotation guidelines to optimize our prompts.
Specifically, for each task, we include two additional instruction components: \ding{182} \textbf{label definitions} and \ding{183} \textbf{annotation notes}. 
For label definition, we provide definitions of all entities for the NER task, and definitions of all relations for the RE task. For the Joint ERE task, which requires the model to perform both NER and RE simultaneously, we provide definitions of both entities and relations. 
For annotation notes, we derive suitable instructions from the human annotation guidelines (see Appendix \ref{apx:guideline}) for each task and provide them to the LLMs. 
We believe that introducing entity and relation definitions and annotation notes offers comprehensive and unambiguous descriptions of the target extraction information. 

In terms of formatting the NER annotation in prompt, we present it as HTML span tag. This is because \cite{wadhwa-etal-2023-revisiting,hu2024improving} demonstrated that when using LLMs for information extraction, the generated results might have the same meaning as in the input text but differs in surface form. 
For example, the entity ``CNNs'' in the input sentence might be generated as ``CNN''. 
To mitigate this error in NER, we instruct the LLMs use HTML span tags to mark all entities in the input sentence to extract the entity spans and use the class attribute to determine the entity types. 
For example, the entity ``CNNs'' in the input text will be marked as ``<span class="Method">CNNs</span>''. 
We provide the complete prompt used in our experiments in Appendix \ref{apx:prompt_design}.

\subsection{Implementation Details} \label{sec:implementation}
For the supervised methods, we use the \textit{scibert-scivocab-uncased} \cite{beltagy-etal-2019-scibert} as encoder.
For the LLMs-based methods, we test the GPT-3.5-Turbo
Llama3-70b, and Qwen2-72b
as the LLM. For few-shot ICL setting, we retrieve 30 demonstrations for each task, and we use the
SimCSE \cite{gao-etal-2021-simcse} as the retriever.
For consistent comparison, all experiments are conducted at the sentence-level.
Appendix \ref{apx:hyper} has  additional implementation details.

\subsection{Evaluation Settings} \label{sec:evaluation}
To evaluate the pipeline extraction and joint ERE, we compute the performance for each subtask, including NER, end-to-end RE (using NER results for relation extraction), and RE (relation extraction with given gold standard entities).
For NER, we conduct span-level evaluation, where both the entity boundary and entity type need to be correctly extracted.
For the end-to-end RE, similar to \cite{zhong-chen-2021-frustratingly, ye-etal-2022-packed,yan-etal-2023-joint}, we report two evaluation metrics: \ding{182} \textbf{Boundaries evaluation} (Rel), which requires the model to correctly predict the boundaries of the subject entity and the object entity, as well as the entity relation; \ding{183} \textbf{Strict evaluation} (Rel+), which further requires the model to predict the entity types based on the requirements of the boundary prediction.
For the RE, given any pair of subject and object entity, the model needs to determine whether a pre-defined relation exists. If a relation does exist, the model must predict the corresponding type.

\section{Experimental Results} 
\subsection{Main Results}
\begin{table*}[!ht]
\centering
\resizebox{0.9\textwidth}{!}{%
\begin{tabular}{lcccccccc}
\toprule
\multicolumn{1}{c}{{ }} & \multicolumn{4}{c}{{ ID Test}} & \multicolumn{4}{c}{{ OOD Test}} \\ \cline{2-9} 
\multicolumn{1}{c}{\multirow{-2}{*}{{ Methods}}} & \multicolumn{1}{l}{{ NER }} & \multicolumn{1}{l}{{ Rel}} & \multicolumn{1}{l}{{ Rel+}} & \multicolumn{1}{l}{{ RE}} & \multicolumn{1}{l}{{ NER}} & \multicolumn{1}{l}{{ Rel}} & \multicolumn{1}{l}{{ Rel+}} & \multicolumn{1}{l}{{ RE}} \\ \hline
\multicolumn{9}{c}{{ \textit{Supervised Baselines}}} \\ \hline
{ PURE \cite{zhong-chen-2021-frustratingly} } & { 81.60} & { 53.27} & { 52.67} & { 73.99} & { 71.99} & { 50.44} & { 49.46} & { 73.63} \\
{ PL-Marker \cite{ye-etal-2022-packed}} & { 83.31} & { 60.06} & { 59.24} & { \textbf{77.11}} & { 73.93} & { 59.02} & { 56.68} & { \textbf{76.83}} \\
{ HGERE \cite{yan-etal-2023-joint}} & { \textbf{86.85}} & { \textbf{62.32}} & { \textbf{61.10}} & { -} & { \textbf{81.32}} & { \textbf{61.31}} & { \textbf{58.32}} & { -} \\ \hline
\multicolumn{9}{c}{{ \textit{Zero-Shot LLMs-based Baselines}}} \\ \hline
{ GPT3.5-Turbo (Joint)} & { 34.76} & { 11.38} & { 10.34} & { -} & { 37.48} & { 10.95} & { 9.97} & { -} \\
{ GPT3.5-Turbo (Pipeline)} & { 51.19} & { 13.57} & { 13.57} & { 35.48} & { 37.73} & { 12.06} & { 11.34} & { 40.74} \\ 
{ Llama3-70b (Joint)} & { 48.87} & { 17.31} & { 17.01} & { -} & { 44.28} & { 17.12} & { 16.63} & { -} \\
{ Llama3-70b (Pipeline)} & { \textbf{61.69}} & { 22.28} & { 21.71} & { 37.35} & { 53.09} & { 27.87} & { 25.57} & { 53.87} \\
{ Qwen2-72b (Joint)} & { 42.15} & { 16.27} & { 14.99} & { -} & { 40.47} & { 15.54} & { 14.31} & { -} \\
{ Qwen2-72b (Pipeline)} & {58.57} & { \textbf{25.76}} & { \textbf{25.76}} & { \textbf{53.50}} & { \textbf{56.43}} & { \textbf{31.25}} & { \textbf{28.13}} & { \textbf{55.37}} \\ \hline
\multicolumn{9}{c}{{ \textit{Few-Shot LLMs-based Baselines}}} \\ \hline
{ GPT3.5-Turbo (Joint)} & { 62.36} & { 23.71} & { 23.49} & { -} & { 51.12} & { 20.12} & { 20.12} & { -} \\
{ GPT3.5-Turbo (Pipeline)} & { 66.27} & { 27.27} & { 24.94} & { 43.26} & { 55.82} & { 22.37} & { 21.49} & { 44.12} \\
{ Llama3-70b (Joint)} & { 63.23} & { 29.21} & { 29.16} & { -} & { 53.12} & { 20.06} & { 19.93} & { -} \\
{ Llama3-70b (Pipeline)} & { \textbf{76.02}} & { 37.55} & { 36.74} & { 56.06} & { \textbf{63.98}} & { 31.33} & { 29.64} & { 62.71} \\ 
{ Qwen2-72b (Joint)} & { 63.73} & { 35.84} & { 34.87} & { -} & { 49.21	} & { 33.17} & { 33.17} & { -} \\
{ Qwen2-72b (Pipeline)} & {71.44} & { \textbf{41.51}} & { \textbf{41.22}} & { \textbf{60.21}} & { 61.72} & { \textbf{39.12}} & { \textbf{37.13}} & { \textbf{63.93}} \\
\bottomrule
\end{tabular}%
}
\caption{Test F1 scores of different baselines on our proposed dataset. ``Joint'' denotes joint ERE. ``Pipeline'' refers to performing NER and RE separately. ``Rel'' and ``Rel+'' denote the results of end-to-end relation extraction under boundaries evaluation and strict evaluation, respectively. ``RE'' indicates performing relation extraction with given gold standard entities, applicable only to pipeline extraction methods. }
\label{tab:main_results}
\vspace{-13pt}
\end{table*}

Table \ref{tab:main_results} reports the experimental results on ID test set and OOD test set.
As described in \S\ref{sec:evaluation}, for the pipeline extraction methods, we present additional RE results when gold standard entities are given.

\noindent\textbf{Supervised Baselines}\quad
We observe that HGERE achieves the best performance on both ID and OOD test sets in NER, Rel, and Rel+, demonstrating the robustness of this current SOTA method.
When comparing the results of ID and OOD, we find that all methods show performance drop on the OOD test set for NER, Rel, and Rel+. 
This is because OOD test provides more challenging and realistic validation scenarios, which require the models to extract information from newly published papers containing new entities. We also observe that the decline in NER scores is more significant, especially for PURE and PL-Marker, whose performance dropped by nearly 10 F1 points. 
This indicates that extracting unseen entities is more challenging for supervised models compared to relation extraction,
which is further supported by the slight decline in RE performance for PURE and PL-Marker in OOD compared to ID. 
We provide a qualitative example in Appendix \ref{apx:error_example}.

\noindent\textbf{LLMs-based Baselines}\quad From the results of both zero-shot and few-shot setting, we have the following observations: \ding{182} Qwen2-72b exhibits the best overall performance than GPT-3.5-turbo and Llama3-70b in both zero-shot and few-shot settings (except the NER task). 
\ding{183} Pipeline extraction outperforms joint ERE in both zero-shot and few-shot settings. Surprisingly, for both Llama3-70b and Qwen2-72b, pipeline extraction shows a significant improvement over joint ERE.
We observed that the NER performance in the pipeline extraction is significantly better than in the joint ERE. 
This indicates that performing LLMs for this end-to-end relation extraction task by decomposing it into seperate NER and RE processes yields better results than joint extraction.
\ding{184} For LLMs-based baselines, the performance of ID does not always outperform OOD and such pattern is very different from supervised baselines.
We believe this is due to the extensive training of LLMs on large-scale data. 
Specifically, for the RE, even though few-shot settings provide similar demonstrations of test data, the ID results are still worse than OOD. 
However, for the NER, Rel, and Rel+ tasks under few-shot settings, the performance on ID tends to be better than on OOD.
Additionally, compared to OOD, the overall performance improvement on ID after using few-shot settings is generally greater than on OOD. This is because, the demonstrations provided to the LLMs are more similar to the ID data.

Previous works \cite{wan-etal-2023-gpt, jimenez-gutierrez-etal-2022-thinking, ma-etal-2023-large} showed that information extraction tasks are very challenging for LLMs compared to supervised methods. 
However, for NER, we found that with appropriate prompt settings, LLMs can be a competent NER model, as reaching an F1 score of 61.69 in zero-shot setting, comparing to the best. 
This suggests that incorporating LLMs into the NER dataset creation process is a feasible solution to reduce human labor.
LLMs perform worse on RE tasks. This is because the test samples for RE tasks contain a large number of NULL labels (see \ref{apx:null_rel}), and large language models have a strong tendency to classify the NULL into predefined types, which has also been confirmed by recent works \cite{jimenez-gutierrez-etal-2022-thinking, wan-etal-2023-gpt}. 
Our experiments show that for end-to-end relation extraction (Rel and Rel+), including the current state-of-the-art (SOTA) models and LLMs-based baselines, there is still significant room for improvement in the future.

\subsection{Ablation Study} \label{sec: ablation}

\begin{figure}[!]
\centering
\includegraphics[width=\linewidth]{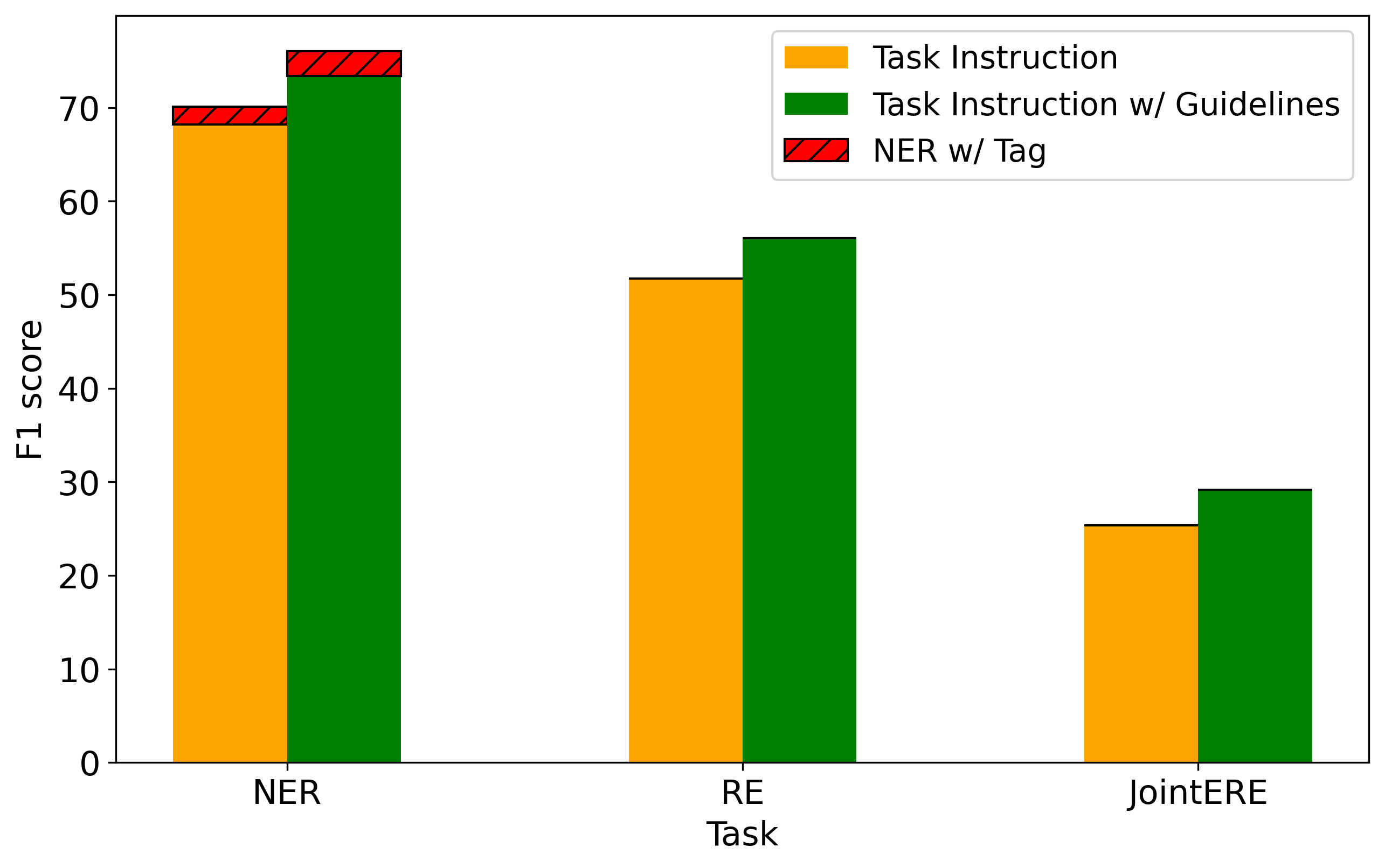}
\vspace{-15pt}
\caption{Ablation study for the effectiveness of using annotation guideline to improve the Instruction ${I}$.
``NER w/ Tag'' denotes the performance gain with additional HTML tag setting. } 
\label{fig:ablation}
\vspace{-10pt}
\end{figure}

To validate the effectiveness of the annotation guideline-enhanced prompt design used in LLM-based baselines, we conducted an ablation study using the Llama3-70b model in a few-shot setting. 
Specifically, for all tasks, we removed the additional instructions derived from the annotation guidelines, retaining only the basic task description in the instruction ${I}$. 
For the NER task, we further removed the requirement of using HTML span tags, allowing the model to directly generate all entities from the input text rather than tagging the input text.
Figure \ref{fig:ablation} presents the results of our ablation study. The results indicate that incorporating label definitions and comprehensive annotation task guidelines significantly improve the model's performance across all tasks. 
Additionally, for NER, the use of HTML span tags further enhances performance.

\subsection{Train Size Experiment}

\begin{figure}[h]
\centering
\includegraphics[width=\linewidth]{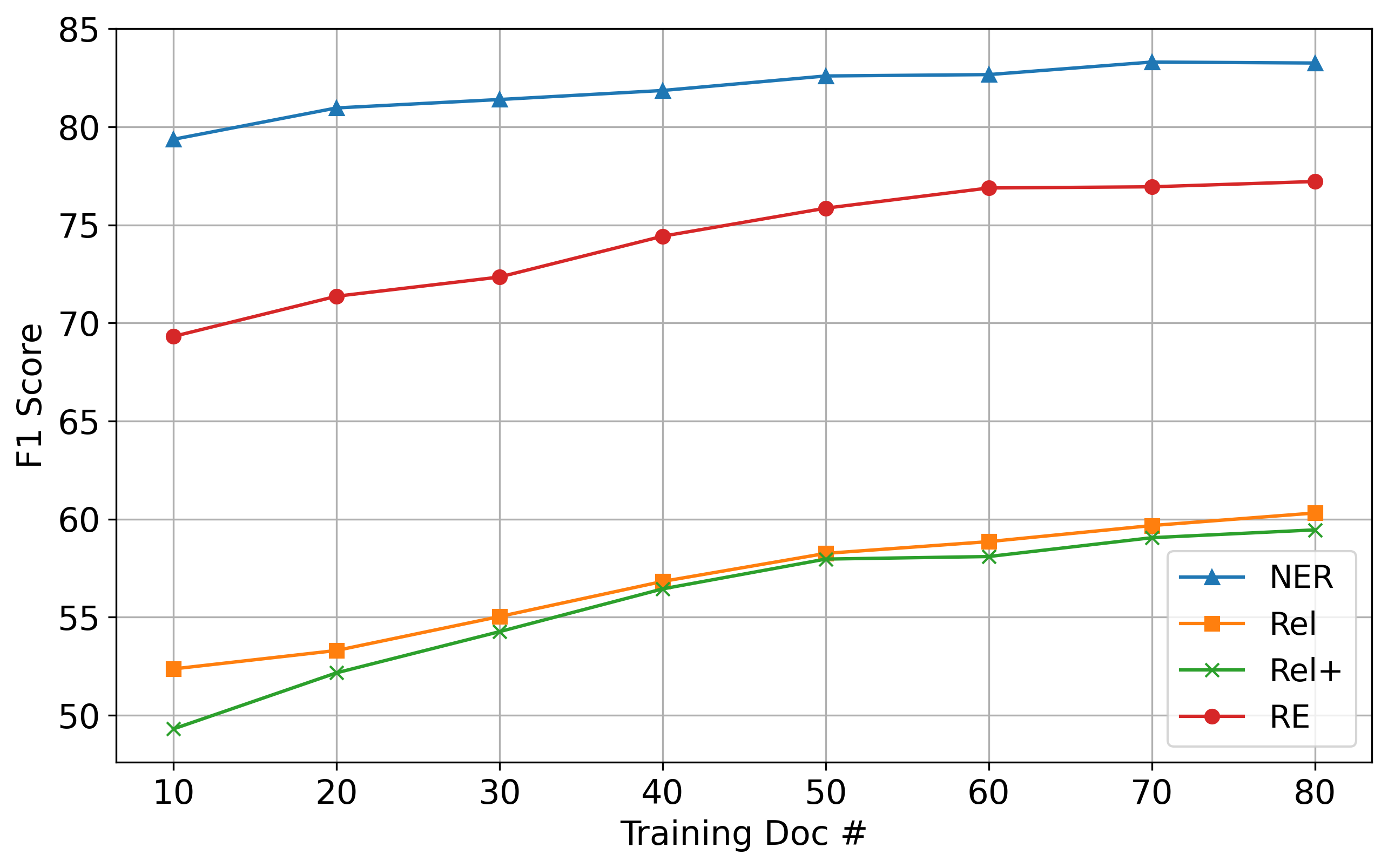}
\vspace{-15pt}
\caption{Performance trends of PL-Marker trained on varying number of documents for NER, end-to-end RE (Rel and Rel+) and RE.} 
\label{fig:train_size}
\vspace{-10pt}
\end{figure}
Annotating datasets for information extraction within specific domains presents certain challenges. Comparing to partial text, such as sentence and abstracts, full-text annotation further exacerbates the difficulties for annotation. 
In the training stage, the number of fully annotated documents plays a crucial role, as documents with fewer annotations have a significant cost advantage. 
We conducted an experiment aimed at assessing the performance of different scientific information extraction tasks across different numbers of training documents.
Figure \ref{fig:train_size} shows the performance trends of the training pipeline extraction model PL-Marker for NER, end-to-end RE (Rel and Rel+), and RE. 
We observe that NER shows a relatively slowed-down improvement as the dataset size increases, suggesting that while it benefits from more data, it experiences diminishing returns when the amount of data becomes large. 
In contrast, both end-to-end RE (Rel and Rel+) and RE show better improvements with an increase in the number of training documents. This indicates that relation extraction is more data-sensitive, requiring more nuanced and varied annotation data for optimal performance.

\section{Conclusion} 
We introduce SciER, a dataset for entity and relation extraction in scientific documents, specifically focusing on datasets, methods, and task entities. To address the limitations of existing datasets, we annotate entire scientific papers for both entities and relations, resulting in a large-scale dataset comprising 106 full-text scientific publications from various AI topics, containing over 24,000 entities and 12,000 relations.
Additionally, we introduce a fine-grained relation set to describe the interactions between datasets, methods, and tasks. 
To evaluate the model's robustness to emporal and conceptual shifts in the SciIE, we also set an OOD test set.

We conduct comprehensive evaluation experiments, including supervised state-of-the-art (SOTA) models and LLM-based ICL baselines, to highlight the challenges in this task. Specifically, for LLM-based methods, we tested both pipeline and joint approaches, optimizing the prompts through retrieval-based ICL, tag-based entity extraction, and the incorporation of annotation guidelines.
The experimental results of LLMs-based methods show that: \ding{182} For the ERE task, pipeline modeling, which decomposes the task into NER and RE sub-tasks, significantly outperforms joint modeling; \ding{183} Although LLM-based approaches require less labeled data, there remains a performance gap compared to supervised methods. 
For future work, we aim to further optimize prompts to enhance the performance of LLMs in Scientific Information Extraction (SciIE) and domain-specific IE tasks. Additionally, a LLM-in-the-loop data annotation system to reduce the high costs of creating domain-specific IE datasets is feasible.

\section*{Limitations}
Despite our diligent efforts, developing a gold standard dataset for entity and relation extraction using a fine-grained and comprehensive relation tag set focused on machine learning datasets, methods, and tasks remains a nontrivial undertaking. This leads to the following limitations associated with the creation of our corpus.
Our dataset only supports three entity types: \texttt{DATASET}, \texttt{METHOD}, and \texttt{TASK}. Incorporating more diverse entity types would be more beneficial for the development of SciIE.
Additionally, many scientific entities are nested, which we have not included.
We also observed that parsing documents from PDF format contains some errors, which increases the difficulty of document processing and cause some of our sentences contain errors.
Finally, we believe that further evluation experiments can be conducted, such as optimizing the ICL baselines for LLMs.
However, due to space constraints, we will consider these as future work.

\section*{Ethical Statement}
The data included in our newly proposed dataset includes a subset of the data collected and freely published by \cite{pan-etal-2024-scidmt-large} within the SciDMT project. All the other data are public from scientific documents.
We release dataset for scientific information extraction tasks. There are no risks in our work.

\section*{Acknowledgements}
This work was supported by the National Science Foundation awards III-2107213, III-2107518, and ITE-2333789. We also thank Saiyun Dong and Faezeh Rajabi Kouchi at Temple University, and Seyedeh Fatemeh Ahmadi at UIC for their valuable contributions to our project.


\clearpage
\appendix

\section{More statistics} \label{apx:dataset_stat}
\subsection{Re-annotating documents from SciDMT} \label{apx:SciDMT_reann}

Table \ref{tab: save_scidmt} presents the details of the entities annotation workload of the 100 documents from SciDMT. 
Specifically, 
the 100 documents from SciDMT-E original contains 21281 entity annotations. 
After our re-annotation process, we compare against the previous SciDMT-E entity annotation, we find that we keep 15989 correctly annotated entities, and remove 709 wrongly annotated entities, fixed 4583 entities and add 2651 new entities. Finally, for this 100 publications we derive from SciDMT contains 23223 entity annotations. Totally, we revived 7234 entities.

\begin{table}[h]
    \centering
    \resizebox{0.9\linewidth}{!}{
    \begin{tabular}{c|cccc|c}
\toprule 
\#Initial & \#Correct & \# Removed  & \#Fixed & \# Added & \# Final  \\ \hline
  21281   & 15989   & 709     & 4583  & 2651  & 23223 \\ \bottomrule
\end{tabular}%
    }
    \caption{The details of our entity annotations efforts for the first 100 documents.}
    \label{tab: save_scidmt}
\end{table}

\vspace{-3pt}
\subsection{Comparison with SciERC}\label{apx:scierc_compare}

Table \ref{tab: scierc_rel_dis} and Table \ref{tab: ent_dist_compare}  show the label statistics of SciERC when only keep the \texttt{DATASET}, \texttt{METHOD}, and \texttt{TASK} entities. We can find that, though SciERC annotated 500 abstract, there are only 1575 entities and 1575 relations related to \texttt{DATASET}, \texttt{METHOD}, and \texttt{TASK}. 

\begin{table}[h]
    \centering
    \resizebox{0.5\linewidth}{!}{
\begin{tabular}{ll}
\toprule
Relation type  & \#   \\ \hline
\texttt{FEATURE-OF}     & 28   \\
\texttt{CONJUNCTION}    & 292  \\
\texttt{USED-FOR}       & 876  \\
\texttt{COMPARE}        & 78   \\
\texttt{HYPONYM-OF}     & 154  \\
\texttt{PART-OF}        & 78   \\
\texttt{EVALUATION-FOR} & 69   \\
Total          & 1575 \\ \bottomrule
\end{tabular}
    }
    \caption{The relation types distribution of datasets (material), methods, and tasks in  SciERC.}
    \label{tab: scierc_rel_dis}
\end{table}

\begin{table}[h]
    \centering
    \resizebox{0.7\linewidth}{!}{
\begin{tabular}{lllll}
\toprule
Dataset & Dataset & Method & Task & Total \\ \hline
SciERC  & 561     & 1592   & 997  & 1575  \\
SciER  & 3942    & 15881  & 4695 & 24518 \\ \bottomrule
\end{tabular}
    }
    \caption{The entity distribution of datasets (material), methods, and tasks in  SciERC and SciER.}
    \label{tab: ent_dist_compare}
\end{table}

\vspace{-5pt}

\vspace{-5pt}
\subsection{SciER Statistics} \label{apx: dataset_stat}

Table \ref{tab: dataset_stat} provide the label distribution of the train, development, ID test and OOD test of our proposed SciER. 

\setlength{\tabcolsep}{3pt}
\begin{table}[h]
    \centering
    \resizebox{0.97\linewidth}{!}{
    \begin{tabular}{lccccc}
\toprule
Rel./Ent. Type      & Train & Dev & ID Test & OOD Test & Total \\ \hline
\texttt{DATASET}              & 11424 & 1549 & 1890     & 1018     & 15881 \\
\texttt{DATASET}               & 3220  & 269  & 370      & 83       & 3942  \\
\texttt{TASK}                 & 3397  & 416  & 688      & 194      & 4695  \\
Total                & 18041 & 2234 & 2948     & 1295     & 24518 \\ \hline
\texttt{PART-OF}        & 1865  & 214 & 304      & 111      & 2494  \\
\texttt{USED-FOR }      & 2398  & 343 & 546      & 167      & 3454  \\
\texttt{EVALUATED-WITH} & 863   & 78  & 131      & 49       & 1121  \\
\texttt{SYNONYM-OF}   & 880   & 76  & 170      & 89       & 1215  \\
\texttt{COMPARE-WITH}   & 875   & 175 & 114      & 54       & 1218  \\
\texttt{SUBCLASS-OF}   & 697   & 114 & 176      & 73       & 1060  \\
\texttt{BENCHMARK-FOR}  & 551   & 64  & 85       & 28       & 728   \\
\texttt{SUBTASK-OF}     & 210   & 31  & 65       & 9       & 315   \\
\texttt{TRAINED-WITH}  & 404   & 37  & 35       & 2       & 478   \\ 
Total          & 8743  & 1132 & 1626     & 582      & 12083 \\
\bottomrule
\end{tabular}
    }
    \caption{The label distribution of our SciER.}
    \label{tab: dataset_stat}
\end{table}
\setlength{\tabcolsep}{6pt}
\vspace{-5pt}

\section{More Implementation Details} \label{apx:hyper}
\vspace{-3pt}
\subsection{Supervised Baselines}
We followed the hyperparameter settings recommended in the PURE, PL-Maker, and HGERE papers respectively. All experiments were conducted using two NVIDIA A100 80GB GPUs for training.
All reported experimental results represent the average of five runs, each with a different random seed.

\begin{table}[h]
    \centering
    \resizebox{\linewidth}{!}{
    \begin{tabular}{lrrr}
    \hline
    Hyperparameter & GPT-3.5-Turbo   & Llama3-70b     & Qwen2-72b        \\ \hline
    Engine         & gpt-3.5-turbo-0125 & Llama3-70b-instruct & Qwen2-72b-instruct  \\
    Temperature    & 0.3                & 0.3          & 0.3        \\
    Max\_tokens    & 256                & 256           & 256       \\
    Top\_p         & 0.9                & 0.9            & 0.9     \\ \hline
    \end{tabular}
    }
    \caption{Hyperparamters of GPT-3.5-turbo, Llama3-70b and Qwen2-72b.}
    \label{tab: gpt_hyper}
\end{table}

\vspace{-8pt}
\subsection{LLM-based Baselines}
The hyperparameters of GPT-3.5-turbo, Llama3-70b, and Qwen2-72b are presented in the Table \ref{tab: gpt_hyper}. 
The used version of SimCSE is \textit{sup-simcse-roberta-large}\footnote{\url{https://huggingface.co/princeton-nlp/sup-simcse-roberta-large}}. 
To ensure fairness in the comparison, we kept the inference hyperparameters consistent for both models. For the GPT-3.5-turbo experiments, due to cost considerations, we sampled 200 sentences from each test set for testing, conducted the tests three times, and then averaged the results. 
The total cost of GPT-3.5-turbo experiments are 50.25 dollars.

For the Llama3-70b and Qwen2-72b, we used two NVIDIA A100 80GB GPUs for inference. 
We tested on all samples in each test set, conducted the tests five times, and then averaged the results.
Due to the input length limitation of Llama3-70b and the lengths of our prompt templates, we set the number of demonstrations for each task as 30, which is also recommended by recent GPT-3 based relation extraction work \cite{wan-etal-2023-gpt}.

\section{More analysis}

\subsection{Qualitative Example} \label{apx:error_example}
\begin{table*}[ht]
\resizebox{\textwidth}{!}{
\centering
\begin{tabular}{lc}
\toprule
 & Example \\
\hline
Ground Truth  & 
\begin{dependency}
\begin{deptext}
Figure 5 shows the process undertaken by \& \textbf{\textcolor{EntMethod}{GxVAEs}} \& for \& \textbf{\textcolor{EntTask}{therapeutic molecular generation}}\&.  \\
\end{deptext}
\depedge[edge height=1ex]{2}{4}{USED-FOR}
\end{dependency}
\\
\hline
PL-Marker  & 
\begin{dependency}
\begin{deptext}
Figure 5 shows the process undertaken by \& \textbf{\textcolor{EntMethod}{GxVAEs}} \& for therapeutic molecular generation.  \\
\end{deptext}
\end{dependency}
\\
\hline
HGERE  & 
\begin{dependency}
\begin{deptext}
Figure 5 shows the process undertaken by \& \textbf{\textcolor{EntMethod}{GxVAEs}} \& for therapeutic \& \textbf{\textcolor{EntTask}{ molecular generation}}\&.  \\
\end{deptext}
\depedge[edge height=1ex]{2}{4}{USED-FOR}
\end{dependency}
\\
\hline
PL-Marker (RE)  & 
\begin{dependency}
\begin{deptext}
Figure 5 shows the process undertaken by \& \textbf{\textcolor{EntMethod}{GxVAEs}} \& for \& \textbf{\textcolor{EntTask}{therapeutic molecular generation}}\&.  \\
\end{deptext}
\depedge[edge height=1ex]{2}{4}{USED-FOR}
\end{dependency}
\\
\hline
Llama3-70b (joint)  & 
\begin{dependency}
\begin{deptext}
Figure 5 shows the process undertaken by \& \textbf{\textcolor{EntMethod}{GxVAEs}} \& for \& \textbf{\textcolor{EntTask}{therapeutic molecular generation}}\&.  \\
\end{deptext}
\depedge[edge height=1ex]{2}{4}{USED-FOR}
\end{dependency}
\\
\hline
GPT-3.5-Turbo(Joint)  & 
\begin{dependency}
\begin{deptext}
Figure 5 shows the process undertaken by \& \textbf{\textcolor{EntMethod}{GxVAEs}} \& for \& \textbf{\textcolor{EntTask}{therapeutic molecular generation}}\&.  \\
\end{deptext}
\depedge[edge height=1ex]{2}{4}{USED-FOR}
\end{dependency}
\\
\hline
Llama3-70b (pipeline)  & 
\begin{dependency}
\begin{deptext}
Figure 5 shows the process undertaken by \& \textbf{\textcolor{EntMethod}{GxVAEs}} \& for \& \textbf{\textcolor{EntTask}{therapeutic molecular generation}}\&.  \\
\end{deptext}
\depedge[edge height=1ex]{2}{4}{USED-FOR}
\end{dependency}
\\
\hline
GPT-3.5-Turbo(pipeline)  & 
\begin{dependency}
\begin{deptext}
Figure 5 shows the process undertaken by \& \textbf{\textcolor{EntMethod}{GxVAEs}} \& for \& \textbf{\textcolor{EntTask}{therapeutic molecular generation}}\&.  \\
\end{deptext}
\depedge[edge height=1ex]{2}{4}{USED-FOR}
\end{dependency}
\\
\bottomrule
\end{tabular}
}
\vspace{-10pt}
\caption{\label{tab:error_example}
Test results of one OOD test example with PL-Marker, HGERE, Llama3-70b (joint), GPT-3.5-Turbo (joint),  Llama3-70b (pipeline), GPT-3.5-Turbo (pipeline). The PL-Marker (RE) means using PL-Marker to predict the relation with given two entities.
}
\end{table*}

Table \ref{tab:error_example} shows one OOD test example for different models. We observe that both PL-Marker and HGERE fail on this example due to the NER results.
PL-Marker ignores the \texttt{TASK} ``therapeutic molecular generation'', and HGERE predicts the wrong span. But if we provide the gold standard entities to PL-Marker, i.e., the PL-Marker (RE) . It predict correctly.
All LLMs-based baselines perform well on this example.

\subsection{Relation Extraction Statistic} \label{apx:null_rel}

We present the proportion of NULL categories in the RE task in the table \ref{tab: null_stat}. We found that the proportion exceeds 60\%.
\begin{table}[t]
    \centering
    \resizebox{0.9\linewidth}{!}{
\begin{tabular}{ccccc}
\toprule
             & \# relation & \# NULL & Tot,  & NULL (\%) \\ \hline
ID test set  & 1626        & 4715    & 6341  & 74.46\%   \\
OOD test set & 582         & 1109    & 1691  & 65.58\%   \\
Dev          & 1132        & 2053    & 3185  & 64.46\%   \\
Train        & 8743        & 20923   & 29666 & 70.53\%   \\ \bottomrule
\end{tabular}
    }
    \vspace{-8pt}
    \caption{Statistics of datasets for relation extraction. ``NULL'' means the given subject and object pairs do not have relation.}
    \label{tab: null_stat}
\end{table}

\vspace{-8pt}
\section{Prompt Design} \label{apx:prompt_design}
\vspace{-3pt}
In this section, we provide the details of annotation guideline-enhanced prompt designs for each task. We list the few-shot version of NER, RE, and Joint ERE. To save the space, we only keep provide 1 demonstration for each task. In our experiments, we use 30 demonstrations. All the zero-shot version are just removed the demonstrations.
\subsection*{Few-Shot NER}
\begin{flushleft}
    \texttt{\textbf{\#\#\# Task:} Generate an HTML version of an input text, marking up specific entities related to machine learning and artificial intelligence. The entities to be identified are: 'Dataset', 'Task', and 'Method'. Use HTML <span> tags to highlight these entities. Each <span> should have a class attribute indicating the type of the entity.
    }
\end{flushleft}
\begin{flushleft}
    \texttt{\textbf{\#\#\# Entity Definitions:} \\- 'Task': A task in machine learning refers to the specific problem or type of problem that a ML/AI model/method is designed to solve. Tasks can be broad, like classification, regression, or clustering, or they can be very specific, such as Pedestrian Detection, Autonomous Driving, Sentiment Analysis, Named Entity Recognition and Relation Extraction...
\\- 'Method': A method entity refers to the approach, algorithm, or technique used to solve a specific task/problem. Methods encompass the computational algorithms, model architectures, and the training procedures that are employed to make predictions or decisions based on data. For example, Convolutional Neural Networks, Dropout, data augmentation, recurrent neural networks...
\\- 'Dataset': A realistic collection of data that is used for training, validating, or testing the algorithms. These datasets can consist of various forms of data such as text, images, videos, or structured data. For example, MNIST, COCO, AGNews, IMDb...
    }
\end{flushleft}
\begin{flushleft}
    \texttt{\textbf{\#\#\# Entity Markup Guide:} \\
    - Use <span class="Task"> to denote a Task entity.
\\
- Use <span class="Method"> to denote a Method entity.
\\- Use <span class="Dataset"> to denote a Dataset entity.
    }
\end{flushleft}

\begin{flushleft}
    \texttt{\textbf{\#\#\# Other Notes:} \\
- Generics cannot be used independently to refer to any specific entities, e.g., 'This task', 'the dataset', and 'a public corpus' are not entities.
\\- The determiners should not be part of an entity span. For example, given span 'the SQuAD v1.1 dataset', where the determiner 'the' should be excluded the entity span. 
\\- If both the full name and the abbreviation are present in the sentence, annotate the abbreviation and its corresponding full name separately. For instance, '20-newsgroup ( 20NG )', the annoation should be '<span class="Dataset">20-newsgroup</span> ( <span class="Dataset">20NG</span> )'.
\\- If one entity with exact same span text appears many times within a sentence, all span text should be marked up.
\\- If one sentence without any entities appear, do not mark up any span text.
\\ - Only annotate “factual, content-bearing” entities. Task, dataset, and method
entities normally have specific names and their meanings are consistent across different
papers. For example, the “CoNLL03”, “SNLI” are factual entities.
\\
- Minimum span principle. Annotators should annotate only the minimum span necessary to represent the original meaning of task/dataset/metric (e.g.: "The", "dataset", "public", ‘method’, ‘technique’ are often omitted).
    }
\end{flushleft}

\begin{flushleft}
    \texttt{\textbf{\#\#\# Examples:} \\
\textbf{Input:} In particular we briefly introduce the principal concepts behind deep Convolutional Neural Networks ( CNNs ) , describe the architectures used in our analysis and the algorithms adopted to train and apply them .
\\ \textbf{Output:} In particular we briefly introduce the principal concepts behind deep <span class="Method">Convolutional Neural Networks</span> ( <span class="Method">CNNs</span> ) , describe the architectures used in our analysis and the algorithms adopted to train and apply them .
}
\end{flushleft}

\begin{flushleft}
    \texttt{\textbf{\#\#\# Input:} Specifically , we investigate the attention and feature extraction mechanisms of state - of - the - art recurrent neural networks and self - attentive architectures for sentiment analysis , entailment and machine translation under adversarial attacks .
}
\end{flushleft}

\begin{flushleft}
    \texttt{\textbf{\#\#\# Output:} 
}
\end{flushleft}

\vspace{-2pt}
\subsection*{Few-Shot RE}
\begin{flushleft}
    \texttt{\textbf{\#\#\# Task:} Based on the given sentence, and subject entity and object entity from the sentence, answer the questions to determine the relationship between them. The potential relations are: ['Part-Of', 'SubClass-Of', 'SubTask-Of', 'Benchmark-For', 'Trained-With',  'Evaluated-With', 'Synonym-Of', 'Used-For' , 'Compare-With']. Answer 'NULL' to indicate that there is no relationship between the entities. 
    }
\end{flushleft}
\begin{flushleft}
\texttt{\textbf{\#\#\# Relationship Definitions:} \\
    - 'Part-Of': This relationship denotes that one method is a component or a part of another method.
\\- 'SubClass-Of': Specifies that one method is a subclass or a specialized version of another method.
\\- 'SubTask-Of': Indicates that one task is a subset or a specific aspect of another broader task.
\\- 'Benchmark-For': Shows that a dataset serves as a standard or benchmark for evaluating the performance of methods on a specific task.
\\- 'Trained-With': Indicates that a method is trained using a specific dataset.
\\- 'Evaluated-With': This relationship denotes that a method is evaluated using a specific dataset to test its performance or conduct the experiments.
\\- 'Synonym-Of': Indicates that two terms or entities are considered to have the same or very similar meaning, such as abbreviation.
\\- 'Used-For': Shows that one entity is utilized for achieving or performing another entity. For example, one Method is Used-For one Task. This relationship is highly flexible, allowing for generic relationships across diverse entities.
\\- 'Compare-With': This relationship is used when one entity is compared with another to highlight differences, similarities, or both.
    }
\end{flushleft}
\begin{flushleft}
    \texttt{\textbf{\#\#\# Notes:} \\
- Determine the 'Relationship' that best describes how the entities are related, or just answer 'NULL' if no relationship exists.
\\- Please do not annotate negative relations. For example, X is not used in Y or X is hard to be applied in Y.
\\- Annotate a relationship only if there is direct evidence or clear implication in the text. Avoid inferring relationships that are not explicitly mentioned or clearly implied.
    }
\end{flushleft}

\begin{flushleft}
    \texttt{\textbf{\#\#\# Examples:} \\
\textbf{Input:} In particular we briefly introduce the principal concepts behind deep Convolutional Neural Networks ( CNNs ) , describe the architectures used in our analysis and the algorithms adopted to train and apply them . 
\\ \textbf{Subject Entity:} Convolutional Neural Network
\\ \textbf{Object Entity:} CNNs
\\ \textbf{Output:} Synonym-Of
}
\end{flushleft}

\begin{flushleft}
    \texttt{\textbf{\#\#\# Input:} Specifically , we investigate the attention and feature extraction mechanisms of state - of - the - art recurrent neural networks and self - attentive architectures for sentiment analysis , entailment and machine translation under adversarial attacks .
    \\ \textbf{Subject Entity:} attention
\\ \textbf{Object Entity:} feature extraction mechanisms
}
\end{flushleft}

\begin{flushleft}
    \texttt{\textbf{\#\#\# Output:} 
}
\end{flushleft}

\subsection*{Few-Shot Joint ERE}
\begin{flushleft}
    \texttt{\textbf{\#\#\# Task:} Identify and extract all relationship triplets consisting of two entities and their relationship from the input text. Each triplet consists of one subject entity, one object entity and their relationship. The interested entity types are: ['Dataset', 'Method', 'Task']. The potential relations are: ['Part-Of', 'SubClass-Of', 'SubTask-Of', 'Benchmark-For', 'Trained-With', 'Evaluated-With', 'Synonym-Of', 'Used-For' , 'Compare-With']. Answer 'NULL' to indicate that there is no triplet.
    }
\end{flushleft}

\begin{flushleft}
    \texttt{\textbf{\#\#\# Entity Definitions:} \\- 'Task': A task in machine learning refers to the specific problem or type of problem that a ML/AI model/method is designed to solve. Tasks can be broad, like classification, regression, or clustering, or they can be very specific, such as Pedestrian Detection, Autonomous Driving, Sentiment Analysis, Named Entity Recognition and Relation Extraction...
\\- 'Method': A method entity refers to the approach, algorithm, or technique used to solve a specific task/problem. Methods encompass the computational algorithms, model architectures, and the training procedures that are employed to make predictions or decisions based on data. For example, Convolutional Neural Networks, Dropout, data augmentation, recurrent neural networks...
\\- 'Dataset': A realistic collection of data that is used for training, validating, or testing the algorithms. These datasets can consist of various forms of data such as text, images, videos, or structured data. For example, MNIST, COCO, AGNews, IMDb...
    }
\end{flushleft}

\begin{flushleft}
\texttt{\textbf{\#\#\# Relationship Definitions:} \\
    - 'Part-Of': This relationship denotes that one method is a component or a part of another method.
\\- 'SubClass-Of': Specifies that one method is a subclass or a specialized version of another method.
\\- 'SubTask-Of': Indicates that one task is a subset or a specific aspect of another broader task.
\\- 'Benchmark-For': Shows that a dataset serves as a standard or benchmark for evaluating the performance of methods on a specific task.
\\- 'Trained-With': Indicates that a method is trained using a specific dataset.
\\- 'Evaluated-With': This relationship denotes that a method is evaluated using a specific dataset to test its performance or conduct the experiments.
\\- 'Synonym-Of': Indicates that two terms or entities are considered to have the same or very similar meaning, such as abbreviation.
\\- 'Used-For': Shows that one entity is utilized for achieving or performing another entity. For example, one Method is Used-For one Task. This relationship is highly flexible, allowing for generic relationships across diverse entities.
\\- 'Compare-With': This relationship is used when one entity is compared with another to highlight differences, similarities, or both.
    }
\end{flushleft}
\begin{flushleft}
    \texttt{\textbf{\#\#\# Notes:} \\
- Input sentence has one triplet: [['entity1 span text:entity1 type', 'relationship', 'entity2 span text:entity2 type']]
\\- Input sentence has no triplets: []
\\- Annotate a relationship only if there is direct evidence or clear implication in the text. Avoid inferring relationships that are not explicitly mentioned or clearly implied.
\\- Ensure that the entity spans are exact extracts from the input text and that the relationships accurately reflect the described interactions. Ensure the output is in the correct format (A list of triplets).
\\- Entities in the triplet should have same form as input sentence.
    }
\end{flushleft}

\begin{flushleft}
    \texttt{\textbf{\#\#\# Examples:} \\
\textbf{Input:} In particular we briefly introduce the principal concepts behind deep Convolutional Neural Networks ( CNNs ) , describe the architectures used in our analysis and the algorithms adopted to train and apply them .
\\ \textbf{Output:} [['CNNs:Method', 'Synonym-Of', 'Convolutional Neural Networks:Method']]
}
\end{flushleft}
\vspace{-5pt}
\begin{flushleft}
    \texttt{\textbf{\#\#\# Input:} Specifically , we investigate the attention and feature extraction mechanisms of state - of - the - art recurrent neural networks and self - attentive architectures for sentiment analysis , entailment and machine translation under adversarial attacks .
    \\ \textbf{Subject Entity:} attention
\\ \textbf{Object Entity:} feature extraction mechanisms
}
\end{flushleft}

\begin{flushleft}
    \texttt{\textbf{\#\#\# Output:} 
}
\end{flushleft}

\vspace{-5pt}
\section{Annotation Guideline} \label{apx:guideline}

This section contains the basic information from our annotation guideline for double-blind review.

\vspace{-5pt}
\subsection{Annotation Tool}

\begin{figure}[t]
\centering
\includegraphics[width=\linewidth]{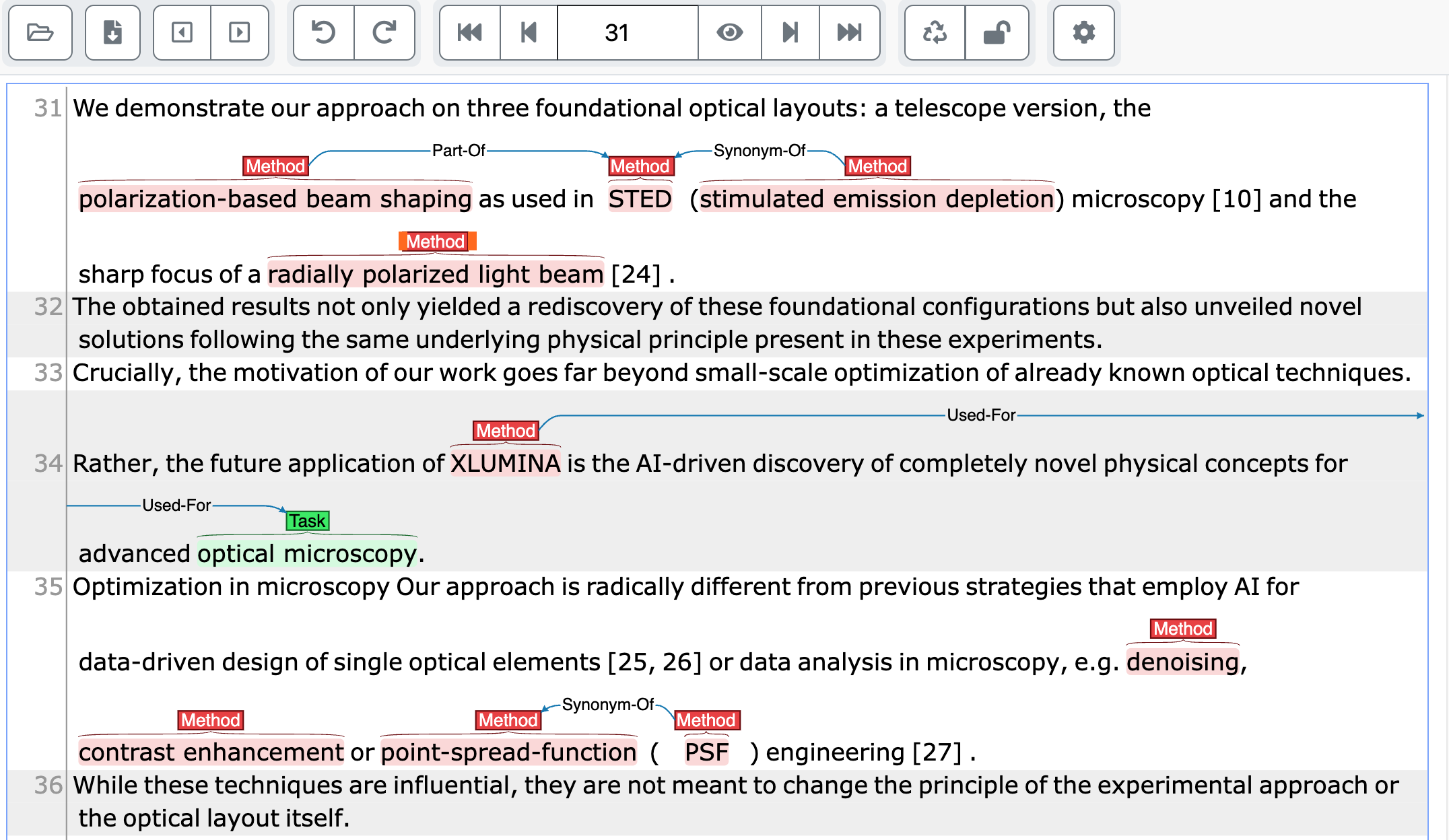}
\vspace{-18pt}
\caption{Annotation interface.} 
\label{fig:interface}
\vspace{-10pt}
\end{figure}

We use the INCEpTION\footnote{\url{https://github.com/inception-project/inception}} as our annotation platform. Figure \ref{fig:interface} shows our annotation interface.

\vspace{-5pt}
\subsection{Entity Annotation}
Scientific entities in the machine learning (ML) or Artificial intelligence (AI) domains refer to key concepts or components that are integral to the structure and study of ML/AI papers. We follow the definition of entities/terms and build our annotation guides for NER based on the ACL RD-TEC Annotation Guideline \cite{qasemizadeh-schumann-2016-acl}, Papers With Code (PwC) and SciDMT \cite{pan-etal-2024-scidmt-large}. We are interested in three specific entity types: Dataset, Task, and Method.

\textbf{Dataset:} A realistic collection of data that is used for training, validating, or testing the algorithms. These datasets can consist of various forms of data such as text, images, videos, or structured data. For example, MNIST, COCO, AGNews, IMDb, etc.

\textbf{Task:} A task in machine learning refers to the specific problem or type of problem that a ML/AI model is designed to solve. Tasks can be broad, like classification, regression, or clustering, or they can be very specific, such as Pedestrian Detection, Autonomous Driving, Sentiment Analysis, Named Entity Recognition and Relation Extraction.

\textbf{Method:} A method entity refers to the approach, algorithm, or technique used to solve a specific task/problem. Methods encompass the computational algorithms, model architectures, and the training procedures that are employed to make predictions or decisions based on data. For example, Convolutional Neural Networks (CNNs), 

\vspace{-3pt}
\begin{flushleft}
\textbf{Annotation Notes:}
\end{flushleft}
\vspace{-3pt}

Considering that annotators may have varying understandings of the annotation details, we have defined a set of rules and notes to standardize the annotation process:

- Do not annotate generics and determiners. Generics cannot be used independently to refer to any specific entities, e.g., “This task”, “the dataset”, “a public corpus” etc. The determiners should not be part of an entity span. For example, given span "the SQuAD v1.1 dataset", where the determiner “the” should be excluded the entity span. We refer ignoring.

- Minimum span principle. Annotators should annotate only the minimum span necessary to represent the original meaning of task/dataset/metric (e.g.: "The", "dataset", "public", ‘method’, ‘technique’ are often omitted).

- Only annotate “factual, content-bearing” entities. Task, dataset, and method
entities normally have specific names and their meanings are consistent across different
papers. For example, the “CoNLL03”, “SNLI” are factual entities.

- If one entity with exact same span text appears many times within a sentence, all span text should be annotated.

\vspace{-5pt}
\subsection{Relation Annotation}

Relation links cannot exceed the sentence boundary. We define 9 types of relations for Dataset, Method, and Task entities.
\vspace{-3pt}
\begin{flushleft}
\textbf{Relation Definitions:}
\end{flushleft}
\vspace{-3pt}

- 'Part-Of': This relationship denotes that one method is a component or a part of another method.

- 'SubClass-Of': Specifies that one method is a subclass or a specialized version of another method.

- 'SubTask-Of': Indicates that one task is a subset or a specific aspect of another broader task.

- 'Benchmark-For': Shows that a dataset serves as a standard or benchmark for evaluating the performance of methods on a specific task.

- 'Trained-With': Indicates that a method is trained using a specific dataset.

- 'Evaluated-With': This relationship denotes that a method is evaluated using a specific dataset to test its performance or conduct the experiments.

- 'Synonym-Of': Indicates that two terms or entities are considered to have the same or very similar meaning, such as abbreviation.

- 'Used-For': Shows that one entity is utilized for achieving or performing another entity. For example, one Method is Used-For one Task. This relationship is highly flexible, allowing for generic relationships across diverse entities.

- 'Compare-With': This relationship is used when one entity is compared with another to highlight differences, similarities, or both.
\vspace{-5pt}
\begin{flushleft}
\textbf{Annotation Notes:}
\end{flushleft}

- Do not annotate negative relations. For example, X is not used in Y or X is hard to be applied in Y.

- Verify that the entities involved in the relation match the prescribed types (e.g., Method-Dataset for Trained-With). Incorrect entity types should not be linked by these specific relations.

- Annotate a relationship only if there is direct evidence or clear implication in the text. Avoid inferring relationships that are not explicitly mentioned or clearly implied.

- Ensure consistency in how relationships are annotated across different texts. If uncertain, refer back to the guideline definitions or consult with a supervisor.

- Do not make assumptions about relationships based on personal knowledge or external information. Rely solely on the information provided in the text.

\end{document}